\documentclass{article} 
\usepackage{iclr2025_conference,times}

\usepackage{amsmath,amsfonts,bm}

\def\Figref#1{Figure~\ref{#1}}

\def\eqref#1{equation~\ref{#1}}

\def\1{\bm{1}}

\def\rvf{{\mathbf{f}}}

\def\rvx{{\mathbf{x}}}

\def\rvz{{\mathbf{z}}}

\def\rmF{{\mathbf{F}}}

\DeclareMathAlphabet{\mathsfit}{\encodingdefault}{\sfdefault}{m}{sl}
\SetMathAlphabet{\mathsfit}{bold}{\encodingdefault}{\sfdefault}{bx}{n}

\def\gL{{\mathcal{L}}}

\definecolor{brightmaroon}{rgb}{0.76, 0.13, 0.28}
\definecolor{brown(web)}{rgb}{0.65, 0.16, 0.16}

\def\btheta{{\boldsymbol{\theta}}}

\usepackage{hyperref}
\usepackage{url}
\usepackage{ulem}
\usepackage{algorithm}
\usepackage{algpseudocode}
\usepackage{graphicx,animate}
\iclrfinalcopy
\usepackage{fancyhdr}
\fancypagestyle{plain}{%
  \fancyhf{}  
  \renewcommand{\headrulewidth}{0pt}  
}
\title{Consistency Training with Physical Constraints}
\author{
Che-Chia Chang$^{1}$, Chen-Yang Dai$^{2}$, Te-Sheng Lin$^{2,3}$, Ming-Chih Lai$^{2}$, Chieh-Hsin Lai$^{2}$ \\
$^1$Institute of Artificial Intelligence Innovation, National Yang Ming Chiao Tung University\\
$^2$Department of Applied Mathematics, National Yang Ming Chiao Tung University \\
$^3$National Center for Theoretical Sciences, National Taiwan University \\
}

\begin{document}

\maketitle

\begin{abstract}
    We propose a physics-aware Consistency Training (CT)~\citep{10.5555/3618408.3619743} method that accelerates sampling in Diffusion Models with physical constraints. Our approach leverages a two-stage strategy: (1) learning the noise-to-data mapping via CT, and (2) incorporating physics constraints as a regularizer. Experiments on toy examples show that our method generates samples in a single step while adhering to the imposed constraints. This approach has the potential to efficiently solve partial differential equations (PDEs) using deep generative modeling.
\end{abstract}

\section{Introduction}

Diffusion models \citep{10.5555/3045118.3045358, NEURIPS2019_3001ef25, ho2020denoising, song2021scorebased} have achieved significant success in high-dimensional data generation. Recent efforts have focused on adapting diffusion models to generate samples that satisfy physical constraints \citep{yuan2023physdiffphysicsguidedhumanmotion, GANTOPO2023, SHU2023111972, jacobsen2024cocogenphysicallyconsistentconditionedscorebased, bastek2024physicsinformeddiffusionmodels}. In physics-informed diffusion models (PIDM) \citep{bastek2024physicsinformeddiffusionmodels}, the authors combine physics-informed neural networks (PINNs) \citep{RAISSI2019686} with diffusion models to sample data distributions while adhering to partial differential equation (PDE) constraints. However, PIDM still suffers from the inherent slow sampling issue of diffusion models. Inspired by PIDM and the recent development of Consistency Training (CT) \citep{10.5555/3618408.3619743}, we propose \textit{CT-Physics}, a method that trains a consistency model with physical constraints from scratch. Unlike PIDM, which requires iterative denoising of samples, CT-Physics generates high-quality samples in one or two steps while ensuring the satisfaction of the system's physical constraints. CT-Physics presents a promising research direction, bridging deep generative models with efficient PDE solving.

\section{Background}

\subsection{Physics-Informed Diffusion Models}

Given $\mathbf{x}_0 \sim q(\mathbf{x}_0)$, we train a diffusion model to iteratively denoise samples while enforcing physical constraints $\bm{\mathcal{R}}(\mathbf{x}_0) = 0$. The model is trained by the denoising loss \(\ell(\btheta)\) and enforce constraints using the residual loss \(\mathcal{R}(\btheta)\) defined as:
\begin{equation*}
    \ell(\btheta) :=  \lambda_t \| \mathbf{x}_0 - \hat{\mathbf{x}}_0(\mathbf{x}_t, t; \btheta) \|^2;\quad \mathcal{R}(\btheta) := \eta_t \| \bm{\mathcal{R}}(\mathbf{x}_0^\text{DDIM}(\mathbf{x}_t, t)) \|^2,
\end{equation*}
where $\lambda_t$ and $\eta_t$ are weights, \(\hat{\mathbf{x}}_0\) is the model estimate of the clean data, and $\mathbf{x}_0^\text{DDIM}$ is the DDIM estimate \citep{song2021denoising} with $N$ steps. The total loss is:
\begin{equation}
    \mathcal{L}_{\text{PIDM}}(\btheta) := \mathbb{E}_{t, \mathbf{x}_0, \rvz \sim \mathcal{N}(\bm{0},\bm{I})} \left[\ell(\btheta) + \mathcal{R}(\btheta)\right].\label{eq:pidm}
\end{equation}

\subsection{Consistency Models}

Consistency Training (CT) enables single-step generation by learning a mapping $\rvf: (\rvx_t, t) \mapsto \rvx_\epsilon$. The function satisfies the self-consistency property
\( \rvf_\btheta(\rvx_t, t) = \rvf_\btheta(\rvx_{t'}, t'), \quad \forall t, t' \in [\epsilon, T], \)
with boundary condition \( \rvf_\btheta(\rvx_\epsilon, \epsilon) = \rvx_\epsilon \). The parameterization ensures the boundary condition: \( \rvf_{\btheta}(\rvx, t) = c_{\text{skip}}(t) \rvx + c_{\text{out}}(t) \rmF_{\btheta}(\rvx), \)
where $\rmF_{\btheta}$ is a neural network and $c_{\text{skip}}(t), c_{\text{out}}(t)$ are differentiable functions with $c_{\text{skip}}(0) = 1$, $c_{\text{out}}(0) = 0$. Self-consistency is enforced via the loss function
\begin{equation*}
    \gL_{\text{CT}}(\btheta) := \mathbb{E}_{t_n, \rvx_0, \rvz} [\ell_{\text{CT}}(\btheta)], \quad \ell_{\text{CT}}(\btheta) := \lambda(t_n) d\left(\rvf_{\btheta}(\rvx_0+t_{n+1}\rvz, t_{n+1}), \rvf_{\operatorname{sg}(\btheta)}(\rvx_0 + t_n \rvz, t_n)\right),\label{eq:loss-ct}
\end{equation*}
where $\lambda(t_n)$ is a weight, $d$ a distance function,
and $\operatorname{sg}(\cdot)$ the stop-gradient operation.

\section{Method and Experiments}
To achieve fast generation while maintaining high sample quality, we propose substituting the diffusion model's less accurate clean prediction, \(\hat{\rvx}_0\), with the consistency model's one-step denoiser, \(\rvf_{\btheta}(\rvx_t, t)\), within the PIDM framework. Consequently, the loss function \(\ell(\btheta)\) in Eq.~\ref{eq:pidm} is replaced by the consistency training loss \(\ell_{\text{CT}}(\btheta)\). To ensure that the samples generated by the consistency model satisfy the physical constraints of the system, we introduce a new loss function given by:
\begin{equation*}
    \mathcal{R}_{\text{CT}}(\btheta) :=\| \bm{\mathcal{R}}(\bm{f}_{\btheta}(\rvx_T, T)) \|^2.
\end{equation*}
Empirically, we found that defining the residual loss with prediction at time \(T\) leads to better enforcement of physical constraints.
The final loss function is defined as:
\begin{equation*}
    \mathcal{L}_{\text{CT-physics}}(\btheta) := \mathbb{E}_{t_n, \rvx_0, \rvz} \left[\ell_{\text{CT}}(\btheta) + \mathcal{R}_{\text{CT}}(\btheta)\right].\label{eq:loss-ct-phy}
\end{equation*}
One might expect that replacing PIDM's clean estimation via the diffusion model with CT would work. However, directly training \(\rvf_\btheta\) using \(\mathcal{L}_{\text{CT-physics}}(\btheta)\) from scratch leads to poor results. We hypothesize that the model overfits the physical constraints, leading to a failure in accurately capturing the original data distribution. See an example in \ref{sec:exp-stage2}.
To address this issue, we propose a two-stage training algorithm:
\begin{enumerate}
    \item Stage 1 (Consistency Training): Train the consistency model only using the consistency loss \(\mathcal{L}_{\text{CT}}(\btheta)\). This stage acts as a warm-up phase, allowing the model to learn the global structure of the data distribution.
    \item Stage 2 (Physics-informed Training): Train the consistency model using the loss function \(\mathcal{L}_{\text{CT-physics}}(\btheta)\), ensuring that the generated samples not only follow the data distribution but also satisfy the physical constraints.
\end{enumerate}
We validate our method using toy examples, with results presented in \Figref{fig:samples}. Additional details and discussions are provided in the appendix.
\begin{figure}[h]
    \centering
    \includegraphics[width=\textwidth]{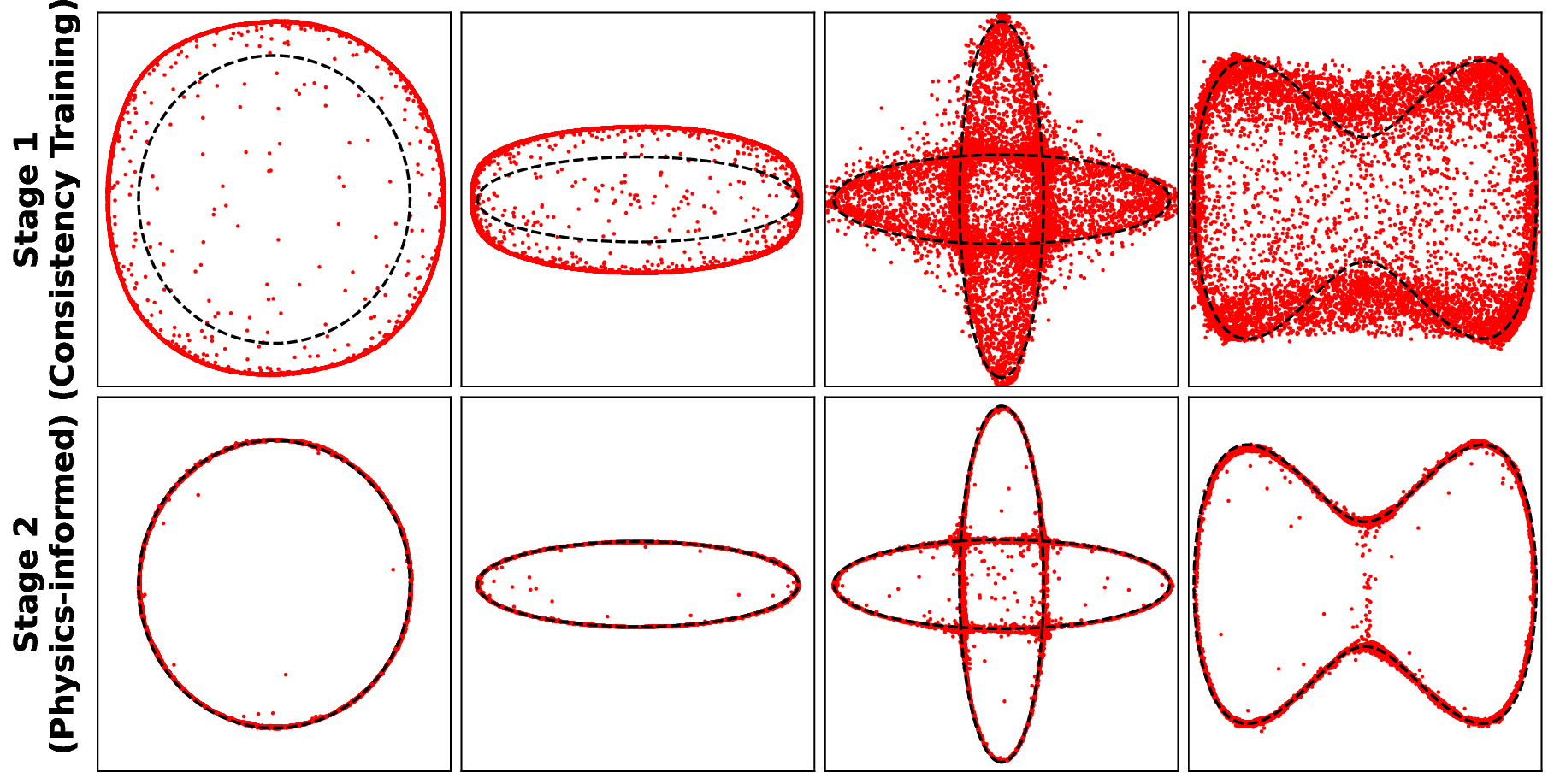}
    \caption{
        Results of CT-Physics on the toy examples. Red dots: model samples, black dashed line: \(\bm{\mathcal{R}}(\rvx_0) = 0\).
    }\label{fig:samples}
    \vspace{-0.5cm}
\end{figure}

\section{Conclusion and Future Work}

We proposed CT-Physics to train consistency models with physical constraints, enabling one-step sampling while ensuring physics constraints are satisfied. Future work includes integrating PDE constraints into training for data generation and efficient PDE solving.


\bibliography{iclr2025_conference}
\bibliographystyle{iclr2025_conference}

\appendix
\section{Appendix}

\subsection{Experimental Details}\label{sec:exp-details}

This section details the experimental setup. For all examples, we will adopt the training settings from improved consistency training (iCT) \citep{song2024improved} except for the maximum number of discretization steps, which is chosen differently for each toy example. For all examples, we sample \(10^4\) data points from the given distribution.

\paragraph{Example 1: Unit Circle}
Let \(\rvx = (x,y)\).
This first example is given by the equation
\begin{equation*}
    \bm{\mathcal{R}}(\rvx) = x^2 + y^2 - 1 = 0.
\end{equation*}
We set the maximum number of discretization steps to 15. The neural network architecture is a 4-layer MLP with 128 hidden units and Sigmoid activation functions. The time step variable \(t\) is transformed into the Fourier feature and then concatenated with the input. We train stage 1 and stage 2 for 1000 epochs, each with a batch size of 128, using the Adam optimizer with a learning rate of \(5 \times 10^{-5}\).


\paragraph{Example 2: Ellipse}

The second example is given by the equation
\begin{equation*}
    \bm{\mathcal{R}}(\rvx) = \frac{x^2}{2^2} + \frac{y^2}{0.5^2} - 1 = 0.
\end{equation*}
The training settings are the same as in the first example.

\paragraph{Example 3: Double Ellipse}

The third example is given by the equation
\begin{equation*}
    \bm{\mathcal{R}}(\rvx) = \left(\frac{x^2}{2^2} + \frac{y^2}{0.5^2} - 1\right)\left(\frac{x^2}{0.5^2} + \frac{y^2}{2^2} - 1\right) = 0.
\end{equation*}
We set the maximum number of discretization steps to 512. The neural network architecture is a 16-layer MLP with 128 hidden units and ReLU activation functions. The time step variable \(t\) is concatenated with the input after sinusoidal embedding. We train stage 1 for 20000 epochs using the RAdam optimizer with a batch size of \(4096\).
The learning rate is set to \(10^{-3}\). We decay the learning rate by half every 1000 iterations to improve numerical stability. For stage 2, we train the model for \(10000\) epochs using the Adam optimizer with a learning rate of \(5 \times 10^{-5}\) with a batch size of \(4096\).

\paragraph{Example 4: Saddle Shape}

The fourth example is given by the equation
\begin{equation*}
    \bm{\mathcal{R}}(\rvx) = x^4 - 2x^2 + y^2 - \frac{1}{4} = 0.
\end{equation*}
We set the maximum number of discretization steps to 256. The neural network architecture is a 4-layer MLP with 128 hidden units and ReLU activation function, with the same time embedding method as in Example 3. We train stage 1 for 10000 epochs using the RAdam optimizer with a batch size of \(512\).
The learning rate is set to \(10^{-3}\). We decay the learning rate by a factor of 0.9 every 1000 iterations for improved numerical stability. For stage 2, we train the model for \(10000\) epochs using the Adam optimizer with a learning rate of \(5 \times 10^{-5}\) with a batch size of \(512\).

\subsection{Stage 2 Training without Stage 1 Warm-up}\label{sec:exp-stage2}

\begin{figure}[h]
    \centering
    \includegraphics[width=0.8\textwidth]{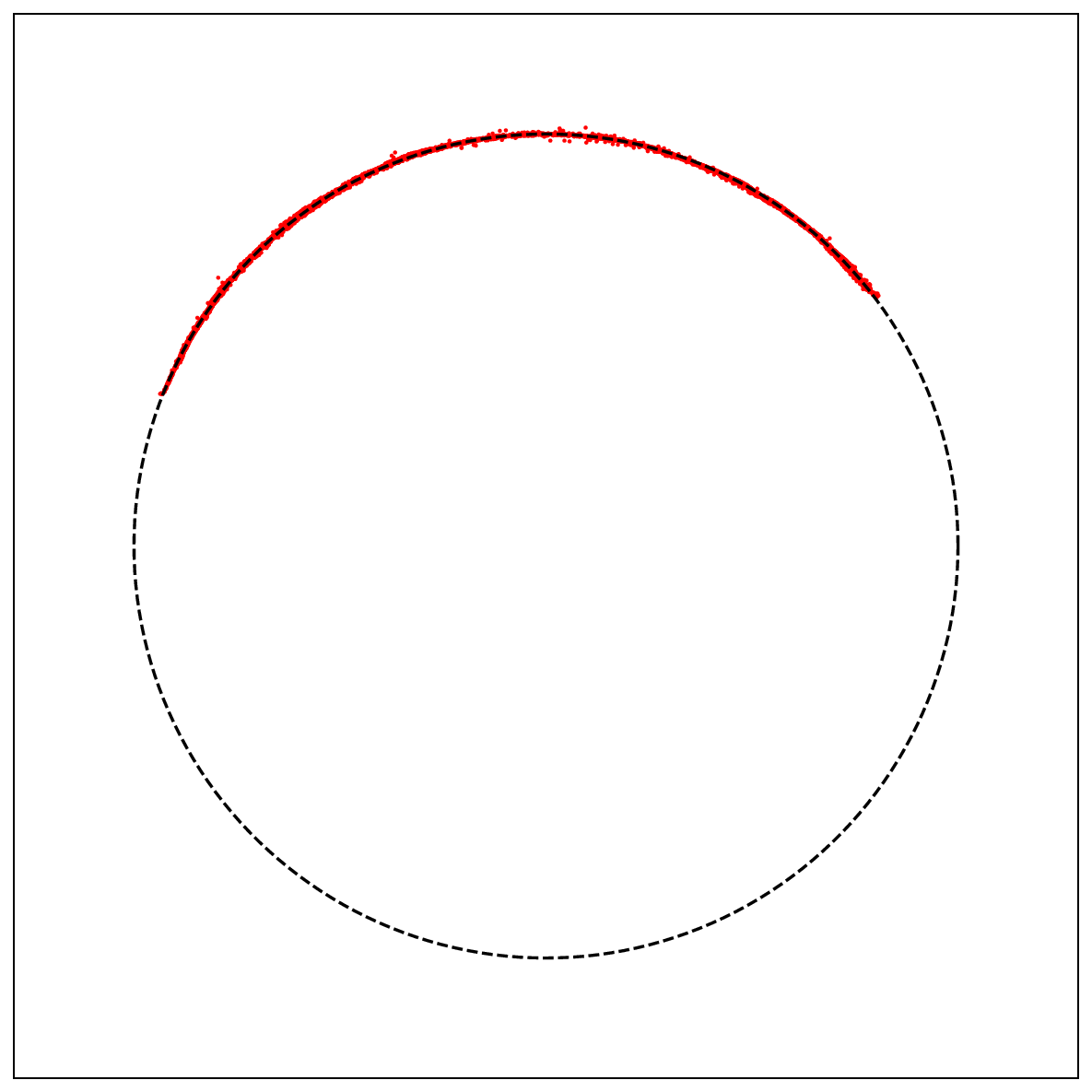}
    \caption{Sampling results of only using Stage 2 training. Red dots: model samples, black dashed line: unit circle. The model fails to capture the original data distribution.
    }\label{fig:sphere_no_teacher}
\end{figure}

We will use the unit circle example from \ref{sec:exp-details} to demonstrate the importance of the warm-up phase. Using the same training settings as in Example 1, we train the model directly from Stage 2 without the warm-up phase. The results are shown in \Figref{fig:sphere_no_teacher}. The model fails to capture the original data distribution, indicating the importance of the warm-up phase in learning the global structure of the data distribution.

\end{document}